%% file: main.tex
\documentclass[11pt, letterpaper, twocolumn]{article} 

\input{structure.tex}

\usepackage{hyperref}
\usepackage{xcolor}
\usepackage[many]{tcolorbox}
\usepackage{balance}
\usepackage{titlesec}
\usepackage{cite}
\definecolor{qagray}{gray}{0.2}
\definecolor{qaline}{rgb}{0.176,0.415,0.619}
\definecolor{syellowcolor}{rgb}{0.98,0.64,0.098}

\newenvironment{QA}{\par\parindent0pt\addvspace{1.0em plus 1pt}\textcolor{syellowcolor}{\color{DarkRed}\vrule width\columnwidth height2pt depth0pt}\vspace*{4pt}\textcolor{qagray}\bgroup\fontsize{11}{13}\selectfont\itshape\leftskip0pt\rightskip0pt plus1fill}{\par\egroup\textcolor{syellowcolor}{\color{DarkRed}\vrule width\columnwidth height2pt depth0pt}\par\addvspace{1.0em plus 1pt}}

\titleformat{\subsubsection}[runin]
  {\normalfont\normalsize\bfseries}{\thesubsubsection}{1em}{}

\title{Interpretation Gaps in LLM-Assisted Comprehension of Privacy Documents}

\author{
	\authorstyle{Rinku Dewri} 
	\newline\newline 
	\institution{Department of Computer Science, University of Denver, CO, 80208, USA}
}

\date{}

\begin{document}

% Preprint IEEE note
\thispagestyle{empty}
\onecolumn

\noindent A full version of this article has now been published by IEEE Computer, available from https://doi.ieeecomputersociety.org/10.1109/MC.2025.3575345. Personal use of this material is permitted.  Permission from IEEE must be obtained for all other uses, in any current or future media, including reprinting/republishing this material for advertising or promotional purposes, creating new collective works, for resale or redistribution to servers or lists, or reuse of any copyrighted component of this work in other works.

\clearpage
\setcounter{page}{1}
\twocolumn

\maketitle

\thispagestyle{firstpage} 

\lettrineabstract{This article explores the gaps that can manifest when using a large language model (LLM) to obtain simplified interpretations of data practices from a complex privacy policy. We exemplify these gaps to showcase issues in accuracy, completeness, clarity and representation, while advocating for continued research to realize an LLM's true potential in revolutionizing privacy management through personal assistants and automated compliance checking.\\}


\begin{figure*}[t]

\centerline{\includegraphics[width=42pc]{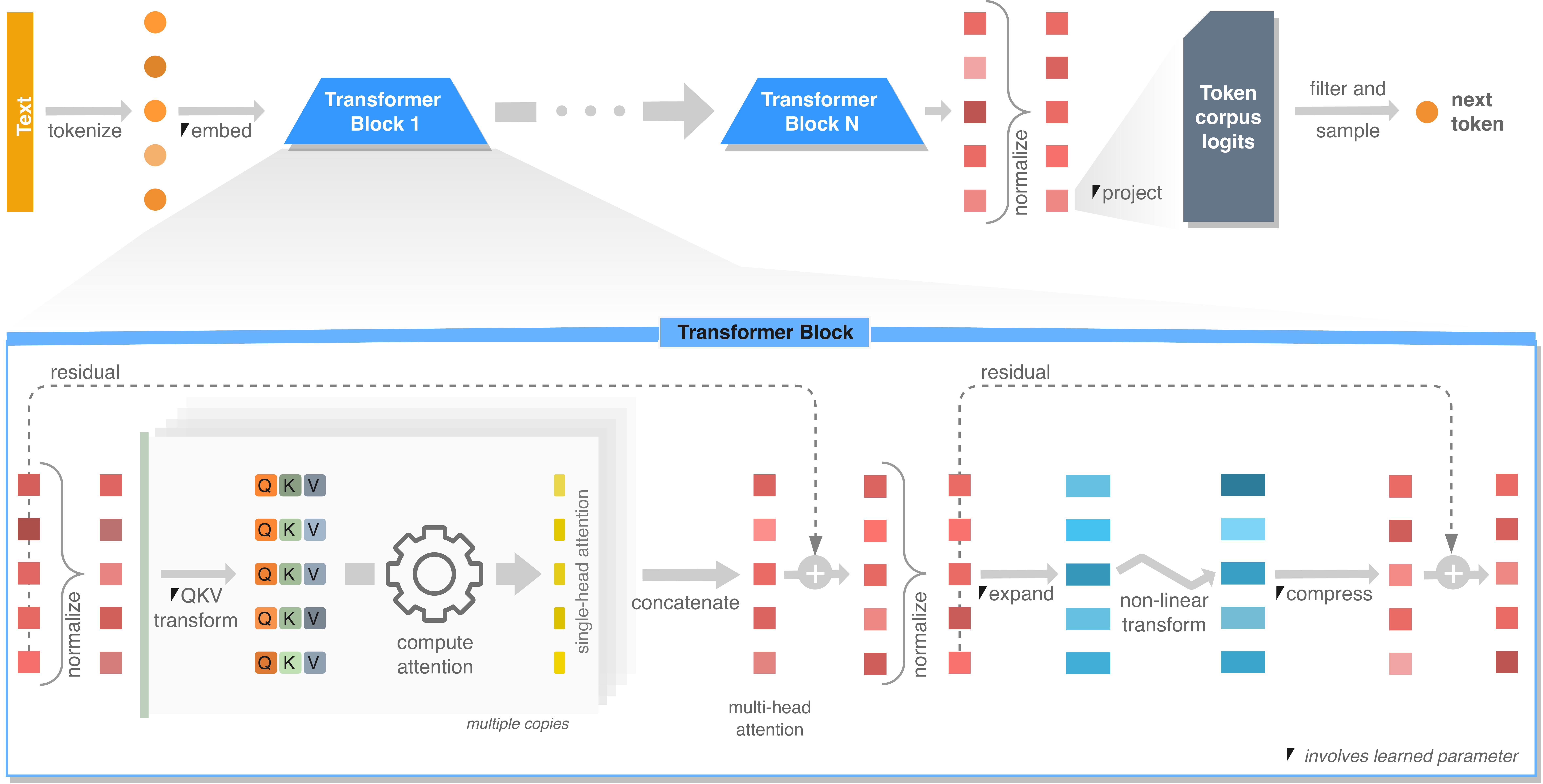}}
\caption{High-level overview of LLM next token generation. The process starts with tokenization and then embedding of the tokens to numerical vectors. The vectors pass through multiple transformer blocks, the final output is normalized (scale adjusted), and then logits (logarithm of odds) are generated for the entire corpus of possible tokens. Logits are converted to probabilities, filtered (say, keep only top-10), and the next token is sampled based on the probability distribution. Each transformer block converts each token to Query-Key-Value (QKV) vectors, computes multi-head attention (to focus on different aspects), and then does expansion-compression to capture non-linear relationships. The process utilizes different parameters learned during training.}\vspace*{-5pt}
\label{fig:llmprediction}
\end{figure*}

\section{INTRODUCTION}

Privacy policies guide us in navigating a complex digital ecosystem, informing what information is gathered, how it is used, and with whom it is shared. However, the reality is far from ideal. Modern privacy policies attempt to address numerous issues, resulting in lengthy documents with regulatory references, technical jargon, and interconnected implications. For many, deciphering such policies has become a barrier to understanding their rights \cite{mcdonald2008}. 

Can language tools ease the burden of comprehending privacy documents? Large language models (LLMs), after their consumer introduction in 2022, are now preferred applications for answering many of our questions. LLMs process and analyze language at scale, and have quickly shown proficiency in summarization, information extraction and question-answering tasks---key elements of a tool that can assist us in privacy policy comprehension.

While LLMs offer exciting possibilities, these tools can misinterpret privacy policies in subtle but significant ways, creating new risks, and potentially exacerbating existing problems. Understanding these risks is crucial to harness the power of LLMs for effective privacy management. In this article, we explore the interpretation errors LLMs may make when processing privacy documents, their causes, and a research-driven path to bridge the gap between current limitations and future potential. 

\section{DOCUMENT COMPLEXITY}
Privacy policies are contractual documents. Data privacy laws do overlap in consumer protection objectives, but also differ in regional or sector-specific philosophies \cite{jordan2022}. For example, while both the General Data Protection Regulation (GDPR) and the California Consumer Privacy Act (CCPA) acknowledge user rights to consent, GDPR defaults to opt-in consent, requiring explicit permission before data collection, while CCPA defaults to opt-out, focusing on the "sale" of personal information. Such differences necessitate verbiage tailored to distinct regulations. This complexity is compounded by the use of legal and technical jargon (e.g., data controller and data processor), conditional phrasing (e.g., if, when), and verb modalities (e.g., may, likely), creating ambiguity and difficulty in determining obligations. 

Privacy policies often reference third-party service providers. Organizations leverage such providers to reduce costs, enhance scalability, and access advanced technologies. However, this practice requires managing multiple data-sharing agreements, leading to dense, layered disclosures. Users must review several interconnected documents to infer their rights and understand how data flows beyond the primary organization.

Beyond the legal and third-party challenges, the structure of privacy policies also adds complexity. The lack of standardization prevents users from adapting to a consistent format. Without standards, information is fragmented, and linked heavily using co-references within the document and cross-references to external documents \cite{adhikari2023}. Policies must also evolve to keep pace with emerging technologies like artificial intelligence (AI), leading to frequent updates, shifting language, and restructuring. 

The effects of these issues impact users and regulators alike. Users are often confused, and ultimately disengage, diminishing the value of privacy policies as a transparency tool. Regulators, struggling to interpret meaning and extract obligations, find it difficult to enforce privacy laws. As natural language policies are likely to stay the norm, innovative solutions are needed to improve comprehension for both users and regulators, facilitating informed consent and effective regulatory oversight.

\section{POLICY COMPREHENSION}
LLMs are emerging as a next-generation tool to interact with information systems through language. LLMs are deep and involved systems trained on massive datasets of text, code and other media to learn complex statistical relationships between words, phrases, sentences and other forms of data modalities. 

\subsection{How LLMs Generate Text?}
Internally, an LLM is a multi-layered neural network with numerous parameters learned during training. During text generation, an LLM first tokenizes sections of text into smaller fragments (e.g., words, subwords or characters), which are then transformed into numerical representations called \textit{embeddings}. The embeddings for a sequence of tokens are processed in subsequent layers of the neural network, ultimately producing a probability distribution over all possible tokens.

A crucial step here is capturing linguistic dependencies and understanding broader contextual relationships. Transformer-based neural network architectures address this using an \textit{attention mechanism} \cite{vaswani2017}, where the relevance of each token is computed in the larger context of a sequence. This mechanism implicitly covers relevance in terms of grammar, meaning, word co-occurrences, and the use of words in a conversation.

With trillions of parameters, and datasets containing hundreds of billions of tokens, LLM training requires massively parallel and distributed computing architectures. Once trained, an LLM can generate human-like text by sampling tokens based on the output probability distribution, enabling applications in natural language processing, content generation, and conversational AI. Figure~\ref{fig:llmprediction} shows a high-level overview of next token prediction in a modern LLM.

\subsection{LLMs for Comprehension}
In the context of a privacy policy, LLMs can leverage their learned embeddings to generate simplified explanations of legal or technical terms, consolidate scattered details to produce cohesive text, and express clauses using alternative wording, potentially surfacing important details buried within complex phrasing. While LLMs are not meant to resolve ambiguous descriptions, they can identify when meaning is not definitive, such as in the use of conditionals, modalities, and vague language (e.g., if, may, likely). Additionally, LLMs can detect when similar clauses in different sections contradict one another \cite{rodriguez2024}. 

Transformer-based LLMs, through the use of their attention mechanism, can systematically retain long contexts when processing text. A \textit{context} here refers to the sequence of tokens that the LLM is weighing to generate new text, which can include an entire privacy document. As a result, LLMs do not have to explicitly track references to other sections in a document, yet they can produce summaries or answer a specific query based on collated information from the entire document. Furthermore, the ability to retain context can be crucial when mapping out data flows to third-party providers, detailing the pre-conditions for such flows, and explaining any user choice mechanism. 

Finally, during training, LLMs are exposed to different patterns and structures through which privacy-related topics are presented. As a result, LLMs implicitly capture the key components present in different policies, and can flag missing information based on standard expectations in privacy documentation. 

 The potential applications of LLMs for privacy comprehension are vast. However, LLMs operate through a computational process that differs from how humans absorb, seek, and generate information, relying on statistical patterns learned from data rather than the intuitive and experiential methods that humans use. 

\section{INTERPRETATION RISKS}
LLMs are driven by context. When a question is posed to an LLM about a privacy-related topic, this context includes not only a privacy document, but also the phrasing of the question, collectively called the \textit{prompt}. Depending on how the question is framed, the LLM prioritizes different pieces of information within the document to generate its response. This reliance on contextual relevance is what makes LLMs powerful, but can also lead to \textit{interpretation gaps}---instances where the LLM’s output deviates from a correct or complete response.

\subsection{Accuracy Issues}
While LLMs are capable of analyzing complex language, they may occasionally generate responses that misstate facts, present incorrect conclusions, or even fabricate plausible-sounding but incorrect information.

\subsubsection*{Misinterpretation}
LLMs can misunderstand the meaning or intent of the policy text when a clause or phrase has a specific legal or technical meaning outside norms. For instance, in GDPR, the term ``data controller'' implies the entity that decides how and why user data is processed. However, in the absence of a GDPR-specific context, the term ``data controller'' may be interpreted as the user who owns the data (as in controls ownership), while ``data processor'' may be interpreted as the organization operating on the data. Misinterpretation is more likely when a privacy document redefines the meaning of established terms, or uses ambiguous language. 

\subsubsection*{Logical Errors}
LLMs may introduce contradictions or illogical statements in their responses. These occur when the model generates conflicting information, makes unwarranted conclusions, or reasons poorly. For example, if a policy clause states that personal data is shared with third-parties, but not sold, depending on the phrasing of the question (shared vs. sold), the LLM can generate seemingly contradicting responses while being technically correct. This issue arises because LLMs are highly sensitive to changes in phrasing \cite{salinas2024}; even small changes can shift attention to different parts of a document.

\subsection{Completeness Issues}
Another significant source of interpretation gaps is related to the completeness of an LLM’s response. In some instances, the model might leave out crucial details or contextually important nuances.

\subsubsection*{Omission}
LLMs may omit essential details that are necessary for a thorough understanding of a policy. For example, they may neglect to include specific user rights, time frames for data retention, or the conditions under which data is shared with third parties, causing misleading conclusions about data rights or organizational obligations. These omissions are often subtle, as the LLM's response may appear complete with details directly relevant to the user's query.

\subsubsection*{Missing Specifics}
A variant of omission is when an LLM does not omit details entirely, but leaves out specific elements. This can include failing to mention particular data types being collected, particular third-party data-sharing arrangements, or specific methods for users to access their data. The omission of these specifics are especially problematic in the legal and regulatory domain, where even minor oversights can lead to significant consequences.

\subsection{Clarity Issues}
Clarity is a critical aspect of privacy policies, and any ambiguity in the interpretation of key clauses or practices may inadvertently introduce vague or imprecise language in a response.

\subsubsection*{Ambiguity}
One of the prevalent issues in privacy policies is ambiguity, which occurs when language is imprecise, overly broad, or undefined. For instance, terms like “reasonable measures” or “best efforts” may be used in a policy without clear definitions. When an LLM processes these vague terms, it may generate responses that retain this ambiguity, further obscuring the meaning. Ambiguity can also arise from legalese sentence structures, which LLMs may fail to simplify sufficiently.

\subsubsection*{Intent vs. Action}
Privacy policies contain a significant number of introductory sentences that provide summative assurances of good privacy practices \cite{adhikari2023}. An LLM may fail to distinguish between such generic statements and actual business practices. For example, if an LLM's response includes language derived from a corporate statement of intent ``\textit{We are committed to adopting state-of-the-art security and privacy technologies to protect our users' data.}" but then lists actual practices that do not align with the intent, the response can be seen as vague, unclear or contradictory.

\subsection{Representation Issues}
Representation issues arise when information is exaggerated, simplified, or incorrectly framed. These errors often occur when prioritizing easy-to-understand summaries over faithfully reflecting policy details.

\subsubsection*{Overstatement}
One common representation issue is overstatement, wherein an LLM exaggerates the control or rights afforded to the user, or inflates a company’s obligations beyond what the original text states. This happens when LLMs simplify complex language, alter weak modalities to strong ones (e.g., might, could to must, will), or omit exception clauses, thereby amplifying the strength of practices. For example, an LLM may overstate the opt-out options, failing to differentiate between data collected for services versus advertising, or falsely suggest guarantees of complete data security while ignoring stated disclaimers.

\subsubsection*{Oversimplification}
An LLM may generate broad statements to improve readability, overly generalizing processes, data flows, or legal requirements, and fail to acknowledge exceptions or limitations. For example, a response may omit exceptions to data retention policies or fail to account for jurisdiction-specific regulations. Oversimplification can also occur when dealing with intricate issues like third-party sharing and international data transfers. For example, an LLM may use the term ``business partners'' without distinguishing between affiliates and advertisers, or summarize a clause about data being stored in ``secure servers'' without noting that these servers are located in a country with weaker data protection laws.

\subsubsection*{False Compliance Assurance}
Privacy policies, being strongly tied to regulatory frameworks, often reference regulations like GDPR or CCPA. An LLM response may include these references from the policy text, even when described practices do not fully comply, or meet only a specific article within a standard. While the training data may include legal text, LLMs generate plausible text (answering the prompt) rather than verifying regulatory alignment. An LLM has to be specifically trained and prompted to carry out such alignment tasks.

\begin{figure}[t]
\begin{tcolorbox}[enhanced, boxrule=0pt, colback=gray!05, drop shadow=black!50!white]
\texttt{I have provided below the text from the IEEE privacy policy. The text begins with the marker ------start and ends with the marker ------end. I will then ask a question. Answer the question.}\\
\\
\texttt{------start}

\texttt{\textbf{[ policy text ]}}

\texttt{------end}\\

\texttt{Question: \textbf{[ question ]}}

\end{tcolorbox}
\caption{Template used to ask questions about the IEEE privacy policy to an LLM. Each prompt contains this entire excerpt, with \textbf{[ policy text ]} replaced by the actual text from the policy and \textbf{[ question ]} replaced by the text of the query.}
\label{fig:prompttemplate}
\end{figure}

\subsubsection*{Data Flow Confusion}
Businesses often involve third-party processors, affiliates, or service providers, and data flows between these entities may not be clearly stated in a policy. Data flow confusion occurs when an LLM fails to distinguish between a company's action on data and that of third parties. For example, the LLM may misidentify who controls or processes the data, omit details about the relationship between the primary organization and third parties, or fail to specify the conditions under which data may be shared. This can blur the lines of responsibility, creating confusion on who handles what data and for what purpose. 

\section{GAPS IN PRACTICE}\label{sec:gaps_in_practice}
To illustrate the real nature of the interpretation gaps, we examine a few concrete cases of LLM errors in practice. We use a Python programming interface to obtain responses from OpenAI's GPT-4o model on the IEEE privacy policy\footnote{Archived version viewable at: \href{https://web.archive.org/web/20250228121024/https://www.ieee.org/security-privacy.html}{https://web.archive.org/ web/20250228121024/https://www.ieee.org/security-privacy.html}}. The full policy text in also provided in the appendix. We asked each question independently using the prompt template in Figure~\ref{fig:prompttemplate}.

\subsubsection*{Example 1: Third-Party Sharing}
IEEE's privacy policy has a paragraph stating that personal data may be shared with third-parties \textit{``in connection with services that these individuals or entities perform for or with IEEE.''} In addition, specific third-party sharing practices are mentioned in other paragraphs. To see how an LLM might collate this information, we posed the following question (response follows next).

\begin{QA}
\lettrine[findent=1pt]{\normalfont W}{hat} information about me and my interactions may be shared by IEEE with third-parties? \\

IEEE may share your personal data with third parties in connection with services that these individuals or entities perform for or with IEEE. This can include the sharing of information for purposes such as completing transactions (e.g., delivery services, credit card clearing houses), hosting an IEEE database, or engaging in data processing on IEEE’s behalf. Additionally, from time to time, IEEE may share your information with third parties whom they believe might provide content, products, or services of interest to you. However, third parties are restricted from using this data in any way other than to provide services for IEEE or for the collaboration in which they and IEEE are contractually engaged.
\end{QA}

Except for the examples relating to completing transactions, much of the text in the response comes from the paragraph specifically discussing third parties. The response lacks specifics on the information shared, and instead focuses on supplemental information surrounding the sharing. For example, the actual policy text includes specifics such as credit card data, usage statistics, conference registration information, and non-personally identifying attributes being shared with one or more of IEEE distributors or resellers, content licensors, advertisers, and subscribing companies and institutions. Changing the question to \textit{``Read each paragraph carefully and list the data items that IEEE may share with third-parties.''} results in a response that includes more of these practices, although not all of them. 

\subsubsection*{Example 2: Opt-Out Choice}
IEEE's privacy policy states that users can access and edit their personal data, and choose to unsubscribe from receiving communications. The policy provides contact information to address issues relating to the use of personal data, but does not mention if users can completely opt-out of data collection. To see if the LLM can distinguish between ``opt-out'' and ``access and edit,'' we posed the following question. 

\begin{QA}
\lettrine[findent=1pt]{\normalfont C}{an} I opt-out of data collection?\\
\\

Yes, you can control the information IEEE has about you and how it is used. If you are a registered user, you can review, revise, and correct the personal data you have provided via IEEE Account/MyProfile. Additionally, you can use the unsubscribe method provided in various communications from IEEE to opt-out of data collection.
\end{QA}

The response indicates a logical error where ``opt-out'' has been misinterpreted to mean the ability to review, revise and correct collected data. The affirming nature of the response is problematic since it confuses the concept of consent withdraw (opt-out) with actively making changes to existing data (edit control). The response also overstates that the unsubscribe method results in opting out of data collection; the policy text only specifies that users can use the unsubscribe method to control how their information is used. 

\subsubsection*{Example 3: Legal Technicality}
IEEE's privacy policy provides a statement that they comply with all applicable data privacy laws and regulations, including the GDPR. Article 17 of GDPR\footnote{Art. 17 GDPR: \href{https://gdpr-info.eu/art-17-gdpr/}{https://gdpr-info.eu/art-17-gdpr/}} details the ``right to erasure,'' under which users have certain rights related to the erasure of their personal data. However, exceptions apply to its enforcement, including ``\textit{for compliance with a legal obligation, or for the establishment, exercise or defense of legal claims.}'' To see if an LLM's response will highlight the interplay between rights and requirements, we asked the following question.

\begin{QA}
    \lettrine[findent=1pt]{\normalfont W}{ill} IEEE comply with my request to erase my data under the GDPR regulation?\\

    Yes, IEEE complies with applicable data privacy laws and regulations, including the General Data Protection Regulation (GDPR). Under these laws, you may have certain rights to your data, which can include the right to request the erasure of your data. If you wish to exercise this right, you should send an email request to privacy@ieee.org with `Data Privacy Request' in the subject line. Additional information may be requested prior to fulfilling a request, and IEEE reserves the right to charge a fee, where permitted, to cover the cost of certain requests.
\end{QA}

The affirmative response creates a false sense of compliance in the beginning, but then switches to ambiguous wordings (``may'' and ``can'') in the following sentence. The remaining response details the procedure to exercise the right, whereas highlighting the possible exceptions could have been more relevant to the question. This response does not imply that the LLM does not have knowledge of the exceptions---adding the sentence \textit{``Provide a comprehensive answer.''} to the question in fact produces a response with a paragraph discussing the exception. 

\begin{QA}
. . . .

However, it is important to note that \textbf{the request to erase personal data is subject to certain conditions and exceptions outlined in the GDPR}. For example, the right to erasure is not absolute and may not apply if the processing of the data is necessary for compliance with a legal obligation, for the performance of a task carried out in the public interest, for the establishment, exercise, or defense of legal claims, among other reasons. 

. . .
\end{QA}

This behavior shows that legal technicalities may not surface in an LLM response, unless some aspect of the question influences the LLM's behavior.

\subsubsection*{Example 4: Data and Purpose Linking}
IEEE's privacy policy lists 18 types of personal data that they collect, which is then followed by 19 paragraphs detailing the different purposes where one or more of these data items may be used. The principle of data minimization in privacy states that organizations should limit data collection only to what is required to serve stated purposes \cite{ISOPDF}. Since policies often do not explicitly provide the association between collected data and purposes, we asked the LLM to create it. 

\begin{QA}
\lettrine[findent=1pt]{\normalfont F}{or} each personal data item that IEEE collects, summarize the purposes stated in the policy.\\

Here is a summary of the purposes for which IEEE collects each personal data item as stated in their privacy policy:

. . .

2. \textbf{Date of birth:} Not explicitly mentioned, but generally related to verifying identity, eligibility for membership, or awards.

. . .

4. \textbf{Emergency contact information:} For ensuring your safety and coordinating assistance if needed.

. . .

6. \textbf{Copies of identification documents:} For verifying identity, compliance with legal/regulatory requirements, and standards development participation documentation.

. . .

16. \textbf{Location information:} For improving relevance of communication and services offered.

. . .

These purposes can overlap for different categories, emphasizing privacy protection, personalization, and improving service delivery.
\end{QA}

The response is accurate for most data items (suppressed above), but has content not directly supported by the policy text. For example, while emergency contact and location information are valid data items that are collected, the policy does not detail how they are used; the text in the LLM's response lists general reasons why such data items are collected. While stating generic purposes can be informative to a user, differentiating them from explicitly stated purposes, as done for ``date of birth,'' is vital to maintaining clarity and correctness in the response. Refining the prompt to say \textit{``For each personal data item that IEEE collects, summarize the purposes \textbf{explicitly} stated in the policy.''} does produce a response where the aforementioned data items are marked as having no stated purpose.

These examples underscore the practical challenges of relying on LLMs for privacy document analysis. Further, they show how LLM responses are sensitive to the wording of a question \cite{salinas2024}, often requiring incorporation of specific words and phrases to extract other contextually relevant information. Repeating similar experiments across multiple policy texts and LLMs is labor-intensive but could provide broader perspectives on the occurrence of these errors.

\section{WHY LLMs STRUGGLE}
Why do even the most sophisticated LLMs struggle with the task of interpreting privacy policies? The answer lies in the inherent complexities of language, compounded by the specific characteristics of how LLMs work. 

\subsubsection*{Pattern Over Accuracy}
Accuracy errors can result when an LLM gives more importance to the general patterns of language observed during training, and overlook specific details in a document. Further, since privacy policies often refer to legal frameworks and industry standards, a thorough response will require an LLM to prioritize such external context over general patterns, a step that LLMs have to be purposefully prompted or trained to do. 

\subsubsection*{Context Constraints}
LLMs often work with a long but fixed-size context. In long conversations, a model may prioritize recent exchanges over earlier inputs, including the policy text itself, leading to responses that are detached from the specifics stated in a policy \cite{liu2024}. Additionally, when an LLM is forced to maintain brevity in responses, details are likely to be lost as the LLM tries to balance information density with readability, and prioritizes commonly relevant details over edge cases \cite{ravaut2024}. 

\subsubsection*{Competing Explanations}
Natural language is inherently ambiguous. Ambiguities in the input (either the question or the policy text) can lead the LLM towards multiple plausible answers with similar statistical probabilities, which can then reflect as multiple interpretations of a practice, or an unclear description if multiple answers are combined into one.

\subsubsection*{Semantic Overlap}
LLMs learn semantics through the use of embeddings and attention. Embeddings capture similarity of words to represent meaning in general discourse. Therefore, words with similar embeddings may be treated as interchangeable, even if they have different connotations in a privacy policy. Similarly, the attention mechanism in LLMs is expected to weigh the relative importance of sections of text. In complex text, the attention mechanism may disproportionately focus on certain sections, even when multiple sections require equal weighting \cite{liu2024}. 

In short, despite having ``seen'' much more data than a human, LLMs are limited by internal mechanisms used to correlate the data. The specific characteristics of privacy policies---complex language and regulatory underpinning---bring such limitations to the forefront.

\section{IMPACT OF ERRORS}
In the mid-1990s, web search changed how we discover information. LLMs take the next step, adding interpretation to existing information, and marking a shift in how we may discover information in future. This new mode of discovery is likely to apply to privacy policies as well, whether due to user preferences, or organizations making LLMs primary communication platforms. Therefore, misinterpretation of policies by LLMs can have widespread consequences for both individuals and organizations.

For individuals, informed choice is essential for trust in a digital ecosystem. Privacy policies, despite their shortcomings, communicate such choices. LLM misinterpretations of these documents can create a false sense of control over one's personal data, leading to misunderstandings of the level of safety offered, unknowing consent to intrusive practices, and finally culminating in an erosion of trust. Misunderstandings can also deter users from adopting safe, valuable services, curbing the growth of innovative data-driven technologies. 

For organizations, LLM errors in interpretation can create legal and financial risks. Large organizations can afford expert reviews of their privacy policies, but smaller organizations may rely on LLMs to analyze policies for compliance with regulations and best practices. LLM misinterpretations of key provisions can put an organization at risk of violating mandatory privacy laws, resulting in significant fines and legal challenges. Additionally, these risks may discourage organizations from adopting the technology to share intelligible overviews of their privacy practices, undermining the potential benefits that LLMs can provide in policy comprehension. 

LLM errors in privacy policy comprehension also raise ethical concerns. The IEEE P7008\textsuperscript{TM} standard highlights how overt or hidden suggestions in intelligent systems can influence user behavior, and pushes for consideration of cognitive biases when designing socio-technical systems. While LLMs offer a path to transparent, user-centric privacy tools, interpretation risks make them prone to misinformation, straying from ethically-responsible and safety-first design principles, and influencing how individuals exercise control over their digital lives. 

\section{THE LLM POTENTIAL}
Despite the risks of misinterpretation, LLMs not only offer a promising avenue to improve policy comprehension, but also open pathways for innovative privacy applications. The complexities of data collection and usage can only be expected to become more intricate in an evolving data-driven ecosystem. Therefore, the design of alternative tools to navigate one's privacy is an impending necessity.

A personal privacy assistant is a technology tool that potentially has a deeper understanding of privacy practices, be it best practices or practices specific to an organization, and uses its understanding to autonomously align user interactions with privacy preferences, proactively inform users of risks, and provide guidance around privacy management \cite{serban2002}. Serving as an interface between a policy and the user, an LLM-based assistant can extract pertinent information from a policy and use it to generate notifications in real time, or compare stated practices with user preferences. Since LLMs are driven by natural language interactions, they will be compelling tools to obtain preferences, generate notices, or detect changes in a policy. 

LLMs can also play a crucial role in automated compliance checking \cite{hassani2024}. The regulatory landscape surrounding data privacy contains multiple laws dictating fair practices within a region, a nation, or across international boundaries. Organizations can use LLMs to analyze their privacy policies and ensure they are in compliance with relevant regulations. This can help businesses identify potential gaps in their policies, strengthen their data protection practices, and reduce the risk of legal challenges. Automated analysis can also help smaller businesses without dedicated legal teams to better understand and implement privacy best practices. These positive impacts are further amplified when frequent regulatory changes are taken into account, along with niche laws applicable to sector-specific businesses. 

While privacy assistants and automated compliance checks are just two examples, we can expect more innovative applications to emerge as these technologies continue to develop. However, realizing this potential requires careful consideration of the challenges and risks discussed earlier, with appropriate safeguards in place to prevent misalignment with the objectives.

\section{THE PATH FORWARD}
Designing reliable AI for privacy document analysis requires continued research, development, and a concerted effort from diverse stakeholders. A key focus is improving LLM accuracy and robustness that can better understand the semantic relationships within policy language, reducing misinterpretation, and reason logically about the implications of different clauses, minimizing logical errors. In addition to larger datasets of legal text and privacy policies, novel algorithms are needed to help an LLM identify associations between policy clauses, privacy impact, and legal premises \cite{lewis2020}. While improved prompting can offer an immediate approach to reduce certain errors, research is also needed to create reasoning models that capture the underlying conceptual structure of privacy and data handling, thereby improving completeness, and enable LLMs to generate more precise and unambiguous language.\cite{fan2024}.

Another area is developing better evaluation metrics. Traditional machine-learning metrics like accuracy are ill-defined for tasks such as summarization or question-answering. Even similarity and semantics-based scoring techniques often fall short \cite{liu2023}. Therefore, new metrics and methods are needed to evaluate LLM responses in a generic setting, but also in ways that are sensitive to the risks of misinterpretation, incompleteness, ambiguity and misrepresentation in privacy communication.

Interdisciplinary collaboration is essential to create a comprehensive privacy tool. The challenges in LLM-assisted privacy comprehension stem from the limitations of technical processes in addressing legal and ethical expectations. AI researchers can continue to address the technical limitations, but the engagement of legal experts and privacy advocates are needed to ensure correct interpretations of privacy law and that technologies respect user autonomy and control.

This will be an iterative process of refinement. In the meantime, adopting an informed approach is crucial to mitigate potential risks. Individuals should not treat LLM responses as definitive interpretations of a privacy policy, but use them for general summaries, look for common pitfalls, and cross-check responses with the original policy. Organizations must ensure human oversight, and adopt structured validation methods when using LLMs to generate policy summaries, revise or restructure a policy, or draft other forms of communication about a privacy practice. It is also important to continuously monitor privacy regulations and ensure that the LLM is not basing outputs on outdated or incorrect information.

\section{CONCLUSION}
LLMs are a ground-breaking technology of this decade. They can be valuable in navigating privacy policies, but current interpretation gaps preclude them from being an authoritative method for policy comprehension. Deficiencies are still being identified and eliminated. Nonetheless, with continued research and development, they can be advanced into a transformative tool for privacy management. This calls for participation of technology researchers, legal experts, privacy advocates and policymakers, and prioritization of collaborative feedback. With a technology such as an LLM in our hands, there are now renewed opportunities for innovation that can bridge the gap between complex policies and their human comprehension.

\def\refname{REFERENCES}

\balance

\newpage
\appendix
\onecolumn

\section*{APPENDIX: Privacy Policy Text}

The following text from IEEE's privacy policy is used for the study in Section~\ref{sec:gaps_in_practice}. The policy has 2807 words. Assuming 100 tokens for 75 words, the prompt fits within the context window limit of GPT-4o API (\href{https://platform.openai.com/docs/models/gpt-4o}{https://platform.openai.com/docs/models/gpt-4o}).\\

\hrule
\vspace{10pt}

\noindent{\large\textbf{IEEE Privacy Policy}}

At IEEE, we respect your privacy. We want to ensure that you get the information, content, and experiences that matter most to you. IEEE is committed to protecting the privacy of its members, customers, volunteers, and other contacts.\\

\noindent{\large\textbf{Scope}}

This privacy policy applies to all personal data processed by full-time and part-time employees, volunteers when acting on behalf of IEEE, contractors and partners doing business on behalf of IEEE, as well as all legal entities, all operating locations in all countries, and all business processes conducted by IEEE.\\

\noindent{\large\textbf{Information collected}}\\

\noindent\textbf{What information do we collect?}

IEEE collects the following personal data in line with the use purposes explained in a subsequent section:

- Your name and contact details

- Date of birth

- Online profile data/usage

- Emergency contact information

- Social media profile information

- Copies of identification documents

- Education and professional information

- Communication information including IEEE Online Support and Contact Center communications

- Purchasing and payment information

- Registration and participation in IEEE events and activities

- Subscription preferences

- Information about the device(s) you use

- Information about service usage

- Cookies

- Authentication data

- Location information

- Author and peer review information

- Other information you upload or provide to us\\

\noindent\textbf{How do we use your information?}

IEEE uses (and, where specified, shares) your personal information for the following purposes:\\

\textit{To process transactions}. IEEE uses personal information such as name, physical address, telephone number, email address, and company/institution to engage in interactions with you, including contacting you about your order, donation, subscription, event participation, or membership. We use financial/credit card and payment information to process your order and may need to share some of this information with delivery services, credit card clearing houses, and other third parties to complete the transaction.\\

\textit{To provide support or other services}. IEEE may use your personal information to provide you with support or other services that you have ordered or requested. IEEE may also use your personal information to respond directly to your requests for information, including registrations for newsletters, webinars, or other specific requests, or pass your contact information to the appropriate IEEE distributor or reseller for further follow-up related to your interests.\\

\textit{To provide information based on your needs and respond to your requests}. IEEE may use your personal information to provide you with notices of new product releases and service developments.\\

\textit{To provide online forums and social networks}. Some services available on the websites permit you to participate in interactive discussions, post comments, opportunities, or other content to a bulletin board or exchange, or otherwise engage in networking activities. Some of these services are moderated; all may be accessed for technical reasons (for example, for improvements or fixes). IEEE does not control the content that users post to these forums or social networks. You should carefully consider whether you wish to submit personal information to these forums or social networks and tailor any content you submit appropriately and in accordance with the relevant terms of use. You should also review any additional terms and conditions that may govern your use of these services, including terms related to sharing your personal information and receiving communications.\\

\textit{To administer products}. IEEE may contact you if you purchase products to confirm certain information about your order (for example, that you did not experience problems in the download process). We may also use this information to confirm compliance with licensing and other terms of use and may share it with your company/institution.\\

\textit{To select content, improve quality, and facilitate use of the websites}. IEEE may use your personal information, including the information gathered as a result of site navigation and electronic protocols and cookies (including third-party cookies), to help create and personalize website content, improve website quality, track marketing campaign responsiveness, evaluate page response rates, conduct usability testing, and facilitate your use of the websites (for example, to facilitate navigation and the login process, avoid duplicate data entry, enhance security, keep track of shopping cart additions, and preserve order information between sessions).\\

\textit{To serve personalized advertising to you}. We don't share your information with advertisers without your consent. We allow advertisers to choose the characteristics of users who will see their advertisements, and we may use any of the non-personally-identifiable attributes we have collected to select the appropriate audience for those advertisements. When you click on or otherwise interact with an advertisement, there is a possibility that the advertiser may place a cookie in your browser and that your attributes meet the criteria the advertiser selected.\\

\textit{To assess usage of IEEE products and services}. IEEE may track your usage of IEEE products and services (e.g., IEEE Xplore) to determine your level of usage, and those usage statistics may be made available to IEEE’s content licensors and your company/institution.\\

\textit{To communicate with you about a meeting, conference, or event}. We may communicate with you about a meeting, conference, or event hosted or co-sponsored by IEEE or one of our business associates. This may include information about the event's content, event logistics, payment, updates, and additional information related to the event. Information you provide when registering for or participating in a conference managed or co-sponsored by parties, other than or in addition to IEEE may be shared with those parties, and the treatment of such information is further subject to the privacy policies of those parties. IEEE and its conference co-sponsors may contact you after the event about the event, subsequent iterations of the event and related events. Please note that IEEE conference, meeting or event organizers, co-sponsors, exhibitors, and other third parties may directly request your personal information at their conference booths or presentations. Providing your information to them is optional, and you should review their privacy policies to address your particular needs and concerns about how they will treat your personal information.\\

\textit{To consider you for a higher member grade or an IEEE award}. IEEE may use personal information such as your name, education, work history, and volunteer activities to consider you for a higher member grade or for an IEEE award. IEEE may need to share this information with your nominators, references, judges, and other parties involved in evaluating your eligibility.\\

\textit{To include you in IEEE Member Directories}. If you are an active member of IEEE, one of its Societies, or IEEE Standards Association, at your request, we may include your information in IEEE Member Directory in IEEE Collabratec, and, if applicable, the Fellows Directory and IEEE Society Directories. We may also enable private member to member messaging (your email address is never shared). Active members can customize their member directories visibility preferences any time in account settings.\\

\textit{To document standards development participation}. If you are a participant in IEEE standards development, IEEE uses your personal information to comply with policies and procedures, legal and accreditation requirements, and evaluation of patent claims by patent offices. IEEE must capture your personal data for these purposes, and provides information about activities related to standards development groups in which you participate. IEEE standards development participation is documented through various methods, e.g., rosters, submission documents, email reflectors, records of meeting attendance, responses to ballots, publicly available participation lists, and declaration of affiliations.\\

\textit{To assist in your participation in IEEE activities}. IEEE will communicate with you, if you are participating in certain IEEE activities such as IEEE conferences, authoring or reviewing an IEEE periodical article, or IEEE humanitarian activities. IEEE may send you information such as newsletters related to those activities.\\

\textit{To update you on relevant IEEE benefits, programs, and opportunities}. IEEE may communicate with you regarding relevant IEEE benefits, programs, and opportunities available to you, through your membership(s) with IEEE.\\

\textit{To connect to 3rd Party Accounts}. On select IEEE sites, you can choose to connect a 3rd party account (e.g. Google Drive) to upload files within the specific site. If you link your 3rd party account, while such connection is active IEEE will have access to the files on such 3rd party account. Such access will include the ability to read file content. IEEE will not share your 3rd party account data with any third parties. IEEE's access to this account will remain until you remove such connection or close your account.\\

\textit{To engage with third parties.} IEEE may share your personal data with third parties in connection with services that these individuals or entities perform for or with IEEE. These third parties are restricted from using this data in any way other than to provide services for IEEE or for the collaboration in which they and IEEE are contractually engaged (for example, hosting an IEEE database or engaging in data processing on IEEE’s behalf, or mailing you information that you requested). These third parties are carefully selected by IEEE and obligated to keep your data secure. From time to time, we may also share your information with third parties whom we think might provide content, products, or services of interest to you.\\

\textit{To protect IEEE content and services}. We may use your information to prevent potentially illegal activities and to enforce our terms and conditions. We also use a variety of technological systems to detect and address anomalous activity and to screen content to prevent abuse, such as spam. These efforts may, on occasion, result in a temporary or permanent suspension or termination of some functions for some users\\

\textit{To get feedback or input from you}. In order to deliver products and services of most interest to our customers, from time to time, we may ask members, customers, volunteers, and website visitors to provide us input and feedback (for example through surveys, usability studies, focus groups).\\

\textit{To protect IEEE information assets as well as your own personal data}. The IEEE Information Security Program protects the confidentiality, integrity, and availability of IEEE information assets by following a risk management approach based on policies, standards, guidelines, and procedures to meet security objectives while supporting business and operational goals.\\

\noindent\textbf{How can you control your information?}

You can control the information we have about you and how we use it in several ways.

If you are a registered user, you can review, revise, and correct the personal data that you have provided to IEEE via IEEE Account/MyProfile. 
    
Use the unsubscribe method provided in our various communications\\

\noindent\textbf{Personal data about minors and children}

IEEE does not knowingly collect data from or about children under 16 without the permission of parent(s)/guardian(s). If we learn that we have collected personal information from a child under 16, we will delete that information as quickly as possible. If you believe that we might have any information from or about a child under age 16, please contact us.\\

\noindent\textbf{How long do we retain your data?}

IEEE will retain your personal information for the length of time needed to fulfill the purposes outlined in this Privacy Policy unless a longer retention period is required or permitted by law.\\

\noindent\textbf{How will you know if the Privacy Policy is changed?}

IEEE may update its Privacy Policy from time to time. If we make any material changes we will notify you by email if you have an IEEE Account, or by means of a notice on this website prior to the change becoming effective. We encourage you to periodically review this page for the latest information on our privacy practices.\\

\noindent{\large\textbf{Technical and regulatory information}}\\

\noindent\textbf{Logging practices}

IEEE automatically records the Internet Protocol (IP) addresses of visitors. The IP address is a unique number assigned to every computer on the internet. Generally, an IP address changes each time you connect to the internet (it is a "dynamic" address). Note, however, that if you have a broadband connection, depending on your individual circumstance, the IP address that we collect may contain information that could be deemed identifiable. This is because, with some broadband connections, your IP address doesn't change (it is "static") and could be associated with your personal computer.

As well as recording the IP addresses of users, IEEE may also keep track of sites that users visited immediately prior to visiting IEEE's website and the search terms they used to find it. We keep track of the pages visited on IEEE's website, the amount of time spent on those pages and the types of searches done on them. Your searches remain confidential and anonymous. IEEE uses this information only for statistical purposes to find out which pages users find most useful and to improve the website.

IEEE also captures and stores information that you transmit. This may include:
\blockquote{
    Browser/Device type/version
    
    Operating system used
    
    Media Access Control (MAC) address
    
    Date and time of the server request
    
    Volume of data transferred
}

\noindent\textbf{External links behavior}

Some of IEEE's websites link to other sites created and maintained by other public- and/or private-sector organizations. IEEE provides these links solely for your information and convenience. When you transfer to an outside website, you are leaving the IEEE domain, and IEEE's information management policies no longer apply. IEEE encourages you to read the privacy statement of each external website that you visit before you provide any personal data.\\

\noindent\textbf{Cookies and web beacons}

Cookies and web beacons are electronic placeholders that are placed on your device by websites to track your individual movements on that website over time. IEEE uses both session-based cookies (which last only for the duration of the user's session) and persistent cookies (which remain on your device and provides information about the session you are in and waits for the next time you use that site again). 

These cookies and web beacons provide useful information to IEEE, enabling us to recognize repeat users, facilitate the user's access to and use of our sites, allows us to track usage behavior, and to balance the usage of our websites on all IEEE web servers. 

Tracking cookies, third-party cookies, and other technologies such as web beacons may be used to process additional information, enable non-core functionalities on the IEEE website and enable third-party functions (such as a social media "share" link). We may also include web beacons and other similar technology in promotional email messages to determine whether the messages have been opened.\\

\noindent\textbf{Do Not Track (DNT)}

The online advertising industry has self-regulatory initiatives designed to provide consumers a choice in the types of ads they may see online and to conveniently opt-out from online behavioral ads served by some or all of the companies participating in these programs. Our websites do not respond to DNT consumer browser settings.\\

\noindent\textbf{Collaboration with authorities}

IEEE has appointed and mandated a privacy officer who represents the regulatory authorities inside the IEEE organization, and in return represents the IEEE organization to regulatory authorities.

The IEEE privacy officer will ensure proper communication with the relevant regulatory authority for privacy. The privacy officer will lead investigative action, complaint handling and data breach notification. The privacy officer will also monitor regulatory changes and consult the regulatory authority where implementation of a regulatory or technological change leads to doubt. \\

\noindent\textbf{Transfer of information to other countries} 

As a global organization, IEEE engages in a number of international activities.  In connection with the management of those activities, IEEE may transfer information to other countries. By submitting your information to IEEE via the websites, or in connection with your interactions with IEEE offline, you consent to such transfers and to the processing of this information in other countries.\\

\noindent\textbf{Responses to legal requests}

IEEE reserves the right to share your information to respond to duly authorized information requests of governmental authorities or where required by law. 

In the event of bankruptcy, merger, acquisition, reorganization or sale of assets, your information may be sold or transferred as part of that transaction. The promises in this privacy policy will apply to your information as transferred to the new entity.

IEEE complies with all applicable data privacy laws and regulations including, but not limited to, the General Data Protection Regulation (GDPR) and the California Consumer Privacy Act (CCPA). Under these laws and regulations, you may have certain rights to your data. Should you wish to exercise any of these rights, please send an email request to privacy@ieee.org with "Data Privacy Request" in the subject line and in the email please identify the specific privacy right you request assistance with. Please note additional information may be requested prior to fulfilling a request and IEEE reserves the right to charge a fee, with respect where permitted, to cover the cost of certain requests.\\

\noindent\textbf{How do I contact you if there is an issue?}

If you have any questions or concerns about this Privacy Policy or about the use of your personal information, please feel free to contact us by email at privacy@ieee.org. \\

\noindent\textbf{Date}

\noindent\textbf{17 January 2025}

\noindent\textbf{Ver 2.0}\\

\noindent\textbf{Contact Information:}

\noindent privacy@ieee.org

\noindent IEEE

\noindent 445 Hoes Lane

\noindent Piscataway, NJ 08854 USA\\

\vspace{10pt}
\hrule

\end{document}

%% file: structure.tex
\usepackage[english]{babel} % English language hyphenation

\usepackage{microtype} % Better typography

\usepackage{amsmath,amsfonts,amsthm} % Math packages for equations

\usepackage[svgnames]{xcolor} % Enabling colors by their 'svgnames'

\usepackage[hang, small, labelfont=bf, up, textfont=it]{caption} % Custom captions under/above tables and figures

\usepackage{booktabs} % Horizontal rules in tables

\usepackage{lastpage} % Used to determine the number of pages in the document (for "Page X of Total")

\usepackage{graphicx} % Required for adding images

\usepackage{enumitem} % Required for customising lists
\setlist{noitemsep} % Remove spacing between bullet/numbered list elements

\usepackage{sectsty} % Enables custom section titles
\allsectionsfont{\usefont{OT1}{phv}{b}{n}} % Change the font of all section commands (Helvetica)

\usepackage{geometry} % Required for adjusting page dimensions

\usepackage[T1]{fontenc} % Output font encoding for international characters
\usepackage[utf8]{inputenc} % Required for inputting international characters

\usepackage{XCharter} % Use the XCharter font

\usepackage{fancyhdr} % Needed to define custom headers/footers
\fancypagestyle{firstpage}{ % Page style for the first page with the title
	\fancyhf{}
}

\newcommand{\authorstyle}[1]{{\large\usefont{OT1}{phv}{b}{n}\color{DarkRed}#1}} % Authors style (Helvetica)

\newcommand{\institution}[1]{{\footnotesize\usefont{OT1}{phv}{m}{sl}\color{Black}#1}} % Institutions style (Helvetica)

\usepackage{titling} % Allows custom title configuration

\newcommand{\HorRule}{\color{DarkGoldenrod}\rule{\linewidth}{1pt}} % Defines the gold horizontal rule around the title

\pretitle{
	\vspace{-30pt} % Move the entire title section up
	\HorRule\vspace{10pt} % Horizontal rule before the title
	\fontsize{26}{30}\usefont{OT1}{phv}{b}{n}\selectfont % Helvetica
	\color{DarkRed} % Text colour for the title and author(s)
}

\posttitle{\par\vskip 15pt} % Whitespace under the title

\preauthor{} % Anything that will appear before \author is printed

\postauthor{ % Anything that will appear after \author is printed
	\vspace{10pt} % Space before the rule
	\par\HorRule % Horizontal rule after the title
	\vspace{0pt} % Space after the title section
}

\usepackage{lettrine} % Package to accentuate the first letter of the text (lettrine)
\usepackage{fix-cm}	% Fixes the height of the lettrine

\newcommand{\initial}[1]{ % Defines the command and style for the lettrine
	\lettrine[lines=3,findent=4pt,nindent=0pt]{% Lettrine takes up 3 lines, the text to the right of it is indented 4pt and further indenting of lines 2+ is stopped
		\color{DarkGoldenrod}% Lettrine colour
		{#1}% The letter
	}{}%
}

\usepackage{xstring} % Required for string manipulation

\newcommand{\lettrineabstract}[1]{\large
	\StrLeft{#1}{1}[\firstletter] % Capture the first letter of the abstract for the lettrine
	\initial{\firstletter}\textbf{\StrGobbleLeft{#1}{1}} % Print the abstract with the first letter as a lettrine and the rest in bold
    \normalsize
}

\usepackage[autostyle=true]{csquotes} % Required to generate language-dependent quotes in the bibliography